\ifx\XeTeXversion\undefined\pdfoutput=1\fi
\documentclass[10pt]{article}
\usepackage[preprint]{tmlr}

\usepackage{amsmath,amsfonts,bm}

\def\eqref#1{equation~\ref{#1}}
\def\1{\bm{1}}

\DeclareMathAlphabet{\mathsfit}{\encodingdefault}{\sfdefault}{m}{sl}
\SetMathAlphabet{\mathsfit}{bold}{\encodingdefault}{\sfdefault}{bx}{n}

\newcommand{\KL}{D_{\mathrm{KL}}}

\usepackage{xcolor}
\usepackage{url}
\usepackage{graphicx}
\usepackage{amsmath}
\usepackage{amssymb}
\usepackage{booktabs}
\usepackage{algorithm}
\usepackage{algpseudocode}
\usepackage{microtype}
\definecolor{citeblue}{HTML}{0F4D92}
\usepackage[
  colorlinks=true,
  linkcolor=citeblue,
  citecolor=citeblue,
  urlcolor=citeblue,
  pdfborder={0 0 0},
  pdfauthor={Keye Zheng, Hanyu Li, Yuan Gao, Zhan Cheng},
  pdftitle={SAVOR: Score Once, Learn Deeper in On-Policy Distillation},
  pdfcreator={},
  pdfproducer={}
]{hyperref}

\title{SAVOR: Score Once, Learn Deeper in On-Policy Distillation}

\author{\name Keye Zheng\textsuperscript{*} \email zhengkeye.zky@alibaba-inc.com \\
      \addr Alibaba Group
      \AND
      \name Hanyu Li\textsuperscript{*} \email hal209@ucsd.edu \\
      \addr University of California, San Diego
      \AND
      \name Yuan Gao\textsuperscript{*} \email ygao31@uw.edu \\
      \addr University of Washington
      \AND
      \name Zhan Cheng \email zhanc113003@gmail.com \\
      \addr University of Wisconsin--Madison
      \AND
      {\normalfont\small \textsuperscript{*}Equal contribution.}}

\def\month{MM}
\def\year{YYYY}
\def\openreview{\url{https://openreview.net/forum?id=XXXX}}

\begin{document}

\maketitle

\begin{abstract}
Standard on-policy distillation (OPD) pays at every step to generate student rollouts and score every response token with a larger teacher, yet discards each scored batch after a single student gradient pass. Simply running additional passes per batch brings limited gains under existing proximal methods because a constant student-teacher mixture weight cannot track a divergence that varies across batches and shrinks over training, either stepping outside a safe trust region on wide-gap batches or under-using teacher supervision as the student improves. We introduce SAVOR, a synchronous OPD method that calibrates target strength per batch to unlock deeper student optimization on already-scored rollouts. Before updating on a batch, SAVOR selects the largest mixture weight inside a sampled reverse-KL and lower-tail reward envelope, freezes the resulting trajectories and token advantages, and updates the student for multiple PPO passes with no additional rollouts or teacher calls. Across five mathematical reasoning benchmarks, SAVOR improves macro accuracy over TOP-D by up to $3.95$ points at a matched teacher-scoring budget and reaches TOP-D's final accuracy using two thirds fewer teacher-scored tokens. Controlled ablations further show that envelope-feasible targets reverse the single-pass advantage of aggressive constant weights under reuse, and that fixing pass depth globally outperforms online token-priced depth selection as update scales grow over training.
\end{abstract}

\section{Introduction}
\label{sec:intro}

On-policy distillation (OPD) transfers reasoning capabilities from a large teacher model to a smaller student by sampling trajectories from the current student policy and querying the teacher for token-level log-probabilities along those visited prefixes~\citep{agarwal2024opd}. Collecting supervision on student-generated states avoids the exposure mismatch of offline sequence distillation~\citep{kim2016sequence,gu2024minillm}, yet carries a steep acquisition cost because every step generates multi-thousand-token student rollouts and runs a dense forward pass of the larger teacher over all sampled tokens. Once that cost has been paid, standard OPD pipelines update the student in a single pass over the batch and immediately discard it. One-pass schedules therefore couple teacher-signal breadth from acquiring new scored rollouts with student realization depth from extracting gradient updates out of trajectories already scored.

Recent trust-region methods refine which teacher signal is acquired on each step. Trust Region Policy Distillation (TOP-D) constructs a proximal target distribution between the student and the teacher and bounds the negative tail of sampled-token rewards so that wide teacher-student gaps remain stable to optimize~\citep{xie2026topd}. Related formulations restrict distillation to high-confidence token regions~\citep{xing2026tropd} or blend rollout policies inside a student-centered KL ball~\citep{plyusov2026trb}. Although TOP-D writes its update rule with a general epoch count, standard configurations execute only a single pass per batch and report aggregate training steps, obscuring whether student performance is bottlenecked by insufficient teacher queries or by shallow optimization of each acquired batch.

\begin{figure}[t]
\begin{center}
\includegraphics[width=\linewidth]{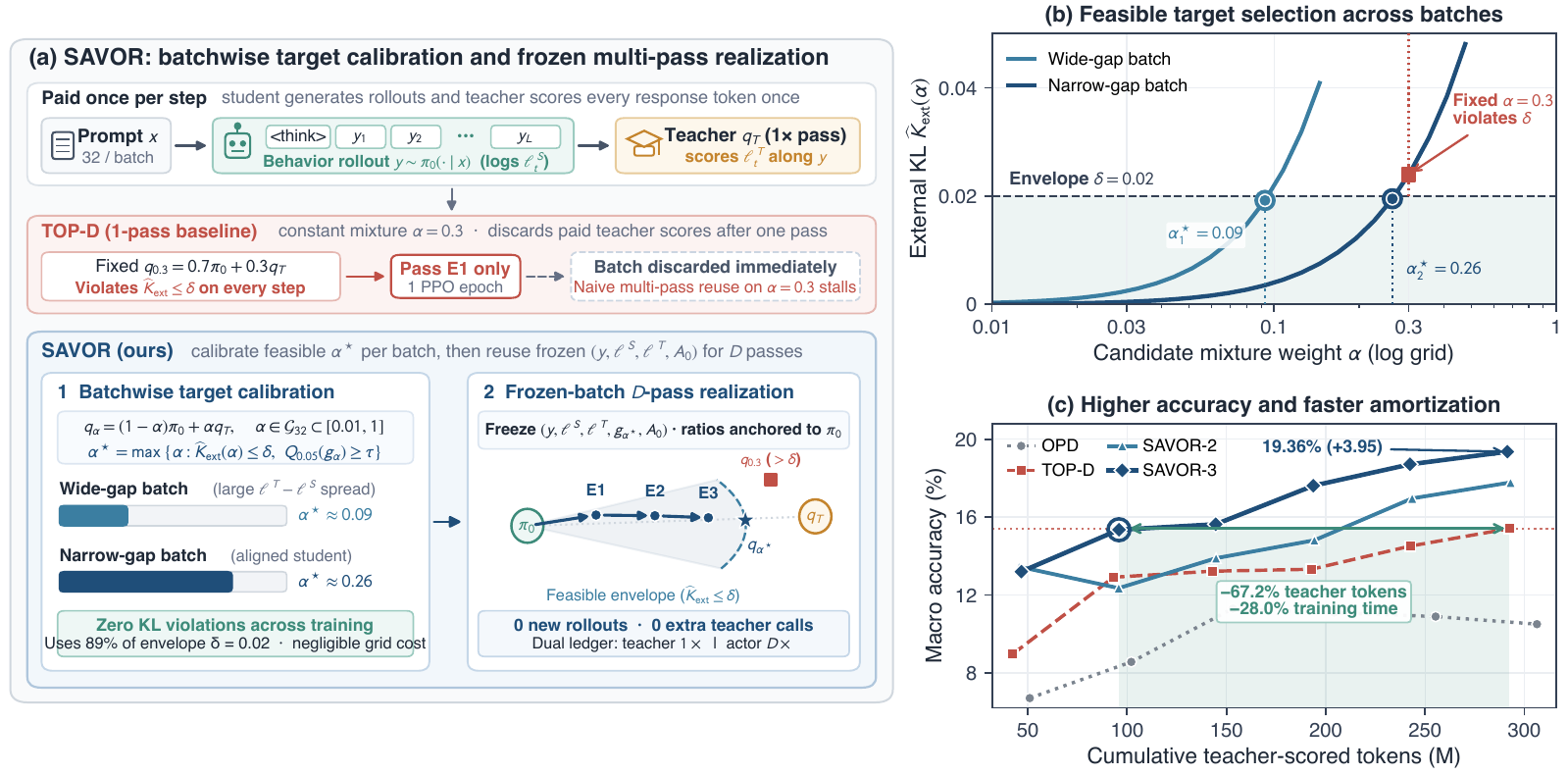}
\end{center}
\caption{Overview of SAVOR. (a) Rollout generation and teacher scoring are paid once per batch. One-pass TOP-D updates the student once under a fixed weight $\alpha=0.3$ that violates the KL envelope on every step and then discards the batch, whereas SAVOR calibrates a feasible $\alpha^\star$ per batch and runs $D$ PPO passes on the frozen tuple $(y, \ell^S, \ell^T, g_{\alpha^\star}, A_0)$ with zero new rollouts or teacher calls. (b) Sampled external reverse KL $\widehat{K}_{\mathrm{ext}}(\alpha)$ over the candidate grid, assigning conservative weights on wide-gap batches and stronger weights on narrow-gap batches inside $\delta=0.02$. (c) Macro accuracy versus cumulative teacher-scored tokens, showing higher final accuracy at matched teacher budgets and faster teacher amortization.}
\label{fig:overview}
\end{figure}

Running multiple PPO passes per batch is a natural remedy~\citep{schulman2017ppo}, yet extra passes alone bring modest gains when the proximal mixture weight is fixed. Because the divergence between teacher and student fluctuates substantially across prompt batches and contracts as the student improves, an aggressive constant weight pulls wide-gap batches outside a local reverse-KL neighborhood while a conservative constant leaves teacher capacity unused on aligned batches. Worse still, tuning $\alpha$ under a single pass actively misleads multi-pass design, as shown in Section~\ref{sec:feasible}. Under one clipped PPO pass, an overly aggressive target shifts probability rapidly on easy tokens while clipping masks immediate damage on outlier prefixes, making the aggressive weight appear superior at one pass even though repeating it drives token ratios into the clipping ceiling and stalls further progress. Multi-pass reuse compounds cleanly only when the target distribution stays feasible on every batch.

Building on this observation, we propose SAVOR, a synchronous OPD method that scores each rollout batch once and learns deeper from it across multiple calibrated passes (Figure~\ref{fig:overview}). Before updating the student on a scored batch, SAVOR searches a log-spaced candidate grid for the largest mixture weight $\alpha^\star$ whose proximal target satisfies both a sampled reverse-KL bound and a lower-tail reward floor. SAVOR then freezes the sampled responses, behavior and teacher log-probabilities, and normalized token advantages, and executes $D$ sequential PPO passes anchored to the rollout policy without new rollouts or teacher calls. Tracking teacher-scored tokens separately from actor-processed tokens allows us to study both fixed multi-pass schedules and online per-batch depth allocation, revealing how growing update scales across training undermine static token-priced depth selection. Our main contributions are three-fold.
\begin{itemize}
    \item \textbf{A realization-depth perspective on on-policy distillation.} We decouple teacher-signal breadth from student optimization depth and show that constant proximal weights cannot track shifting teacher-student divergence across batches, causing single-pass tuning to favor out-of-envelope targets that stall under reuse.
    \item \textbf{Batchwise target calibration for frozen multi-pass updates.} We propose SAVOR, which selects the strongest mixture target inside a sampled reverse-KL and reward-tail envelope for each batch and optimizes the student for multiple PPO passes over frozen trajectories and advantages with zero extra rollouts or teacher calls.
    \item \textbf{Main empirical results and controlled ablations.} Distilling Qwen3-30B-A3B into Qwen3-1.7B-Base across five mathematical reasoning benchmarks, SAVOR improves macro accuracy over TOP-D by up to $3.95$ points at matched teacher tokens and matches TOP-D's final accuracy while cutting teacher scoring by two thirds. Controlled ablations isolate why multi-pass gains require both feasible targets and frozen advantages, and explain why fixed global pass depth outperforms pilot-calibrated token pricing.
\end{itemize}

\section{Method}
\label{sec:method}

\subsection{Preliminaries and proximal teacher rewards}
\label{sec:prelim}

For a prompt $x$, let $y=(y_1,\ldots,y_L)$ denote a response of $L$ valid tokens sampled from the current student behavior policy $\pi_0(\cdot\mid x)$. At token position $t$, write $\ell^S_t=\log\pi_0(y_t\mid x,y_{<t})$ and $\ell^T_t=\log q_T(y_t\mid x,y_{<t})$ for the behavior student and teacher log-probabilities. Standard OPD~\citep{agarwal2024opd} minimizes reverse KL along student trajectories using the sampled-token log-ratio reward
\begin{equation}
    r^{\mathrm{OPD}}_t = \ell^T_t - \ell^S_t.
    \label{eq:opd}
\end{equation}
When $q_T$ assigns tiny probability to tokens sampled by a weaker student, $r^{\mathrm{OPD}}_t$ suffers from sharp negative spikes. TOP-D~\citep{xie2026topd} bounds those spikes by interpolating a token-level proximal target $q_\alpha(\cdot\mid x,y_{<t})=(1-\alpha)\pi_0(\cdot\mid x,y_{<t})+\alpha q_T(\cdot\mid x,y_{<t})$ with constant $\alpha\in(0,1)$. Its sampled-token log-ratio reward is
\begin{equation}
    g_{\alpha,t} = \log\left(1 - \alpha + \alpha e^{\ell^T_t - \ell^S_t}\right),
    \label{eq:topd}
\end{equation}
which is bounded below by $\log(1-\alpha)$ and approaches zero as $\alpha\to 0$. TOP-D converts $g_{\alpha,t}$ into a position-averaged future token return
\begin{equation}
    R_t = g_{\alpha,t} + \frac{\sum_{j=t+1}^{L} g_{\alpha,j}}{\max(L-t,1)}
    \label{eq:return}
\end{equation}
and normalizes $R_t$ within each prompt group to form the pre-update token advantage $A_0$. Whereas TOP-D divides centered returns by each prompt group's standard deviation, SAVOR preserves cross-prompt return scale by dividing by a slow exponential moving average of median absolute centered returns across batches. Although TOP-D writes $E$ internal iterations in general form, its experiments fix $E=1$. Appendix~\ref{app:normalization} gives both normalization formulas.

\subsection{Selecting the strongest feasible target per batch}
\label{sec:target}

Because the divergence between $q_T$ and $\pi_0$ shifts across prompt batches and shrinks over training, no single constant $\alpha$ maintains a consistent trust-region step. Large-gap batches are pulled far outside a local neighborhood of $\pi_0$ under an aggressive constant, while small-gap batches under-use teacher supervision under a conservative one.

SAVOR adapts $\alpha$ to each batch prior to student optimization, forming a two-sided trust region in which acquisition bounds the external step from $\pi_0$ to $q_{\alpha^\star}$ and clipped PPO bounds the internal step from $\pi_0$ to the updated student. Over a 32-point log-spaced grid $\mathcal{G}\subset[0.01,1]$, SAVOR selects the largest $\alpha^\star\in\mathcal{G}$ satisfying
\begin{equation}
  \begin{aligned}
    \widehat{K}_{\mathrm{ext}}(\alpha)
    &= \left\langle e^{g_\alpha} - 1 - g_\alpha \right\rangle_m \leq \delta, \\
    Q_{0.05}(g_\alpha) &\geq \tau,
  \end{aligned}
  \label{eq:alpha}
\end{equation}
where $\langle\cdot\rangle_m$ averages over valid response tokens in the batch and $Q_{0.05}(g_\alpha)$ is their fifth percentile. Because $e^{g_{\alpha,t}} = q_\alpha(y_t\mid x,y_{<t})/\pi_0(y_t\mid x,y_{<t})$ on tokens sampled from $\pi_0$, the pointwise nonnegative function $k_3(u)=e^u-1-u\ge 0$ is the low-variance Bregman estimator of the sampled-path reverse KL divergence $\KL(\pi_0 \Vert q_\alpha)$~\citep{schulman2017ppo}. We call $\widehat{K}_{\mathrm{ext}}(\alpha)$ the external K3 divergence to separate this pre-update target gap from post-update student drift. The quantile floor $Q_{0.05}(g_\alpha)\ge\tau$ guards against heavy negative reward tails on outlier prefixes. If no grid point satisfies both bounds, the solver defaults to $\min\mathcal{G}$.

Evaluating Equation~\ref{eq:alpha} over 32 scalar candidates reuses cached log-ratios $\ell^T_t-\ell^S_t$ and adds negligible compute. Batches with a wide teacher-student gap receive a smaller $\alpha^\star$ to stay within $\delta$, whereas batches with a narrow gap receive a larger $\alpha^\star$ so the teacher signal remains as strong as the envelope permits.

\begin{algorithm}[t]
\caption{One SAVOR training step}
\label{alg:savor}
\begin{algorithmic}[1]
\Require Student policy $\pi_\theta$, teacher $q_T$, prompt batch $X$, pass depth $D$, KL envelope $\delta$, tail floor $\tau$, candidate grid $\mathcal{G}$
\State $\pi_0 \gets \pi_\theta$
\State Sample rollout responses $y \sim \pi_0(\cdot\mid x)$ for all prompts $x \in X$ and record $\ell^S_t = \log\pi_0(y_t\mid x,y_{<t})$
\State Query teacher $q_T$ once on $(x, y)$ to obtain token log-probabilities $\ell^T_t = \log q_T(y_t\mid x,y_{<t})$
\State Select $\alpha^\star \gets \max\{\alpha \in \mathcal{G} : \widehat{K}_{\mathrm{ext}}(\alpha) \le \delta \text{ and } Q_{0.05}(g_\alpha) \ge \tau\}$ via Equation~\ref{eq:alpha}
\State Compute proximal rewards $g_{\alpha^\star,t}$, token returns $R_t$, and normalized advantages $A_0$
\State Freeze batch tuple $(X, y, \ell^S, \ell^T, g_{\alpha^\star}, A_0)$ and valid-token masks
\For{pass $k = 1$ to $D$}
  \State Update $\pi_\theta$ for one PPO epoch of mini-batches on the frozen tuple maximizing Equation~\ref{eq:ppo}
\EndFor
\end{algorithmic}
\end{algorithm}

\subsection{Reusing the frozen batch for a fixed number of passes}
\label{sec:reuse}

Once $q_T$ has scored the batch and $\alpha^\star$ is fixed, the tuple $(X, y, \ell^S, \ell^T, g_{\alpha^\star}, A_0)$ forms a calibrated supervision asset. SAVOR freezes this tuple and runs $D$ complete actor passes, labeled E1 through E$D$, producing $\pi_1,\ldots,\pi_D$. Each pass sweeps the frozen batch in one epoch of shuffled mini-batches maximizing the clipped PPO surrogate~\citep{schulman2017ppo}
\begin{equation}
  \mathcal{J}_{\mathrm{PPO}}(\theta) = \left\langle \min\!\left(r_t(\theta) A_{0,t},\; \operatorname{clip}\!\left(r_t(\theta),\, 1-\epsilon_{\mathrm{clip}},\, 1+\epsilon_{\mathrm{clip}}\right) A_{0,t}\right) \right\rangle_m,
  \quad r_t(\theta) = e^{\log\pi_\theta(y_t\mid x,y_{<t}) - \ell^S_t},
  \label{eq:ppo}
\end{equation}
with all ratios $r_t(\theta)$ anchored to $\pi_0$. Passes E2 through E$D$ require no new rollouts and no teacher calls (Algorithm~\ref{alg:savor}). We denote the deterministic $D=2$ and $D=3$ schedules by SAVOR-2 and SAVOR-3.

Two invariants govern reuse. First, $A_0$ is never recomputed or reweighted between passes, because redefining residual rewards after E1 alters the optimization target mid-step and degrades accuracy as verified in Section~\ref{sec:feasible}. Second, SAVOR logs teacher-scored tokens once upon acquisition and actor-processed tokens once per executed pass, separating teacher breadth from student realization depth.

\subsection{Pass-level diagnostics and fixed versus adaptive depth}
\label{sec:depth_theory}

Comparing SAVOR with multi-epoch PPO~\citep{schulman2017ppo}, GRPO~\citep{shao2024deepseekmath}, and TOP-D~\citep{xie2026topd} raises a natural design question of whether pass depth $D$ should be fixed globally or chosen per batch. To measure each pass, we re-forward $\pi_k$ on the frozen batch after pass $k\in\{1,\ldots,D\}$ to obtain cumulative displacement $\Delta^{(k)}_t = \log\pi_k(y_t\mid x,y_{<t}) - \ell^S_t$ and ratio $\rho^{(k)}_t = \exp(\Delta^{(k)}_t)$. Internal student reverse-KL drift $\KL(\pi_0\Vert\pi_k)$ and normalized effective sample size (ESS) are estimated by
\begin{equation}
  \widehat{K}_{\mathrm{int}}^{(k)} = \left\langle \rho^{(k)} - 1 - \Delta^{(k)} \right\rangle_m,
  \qquad
  \mathrm{ESS}^{(k)} = \frac{\langle \rho^{(k)} \rangle_m^2}{\langle (\rho^{(k)})^2 \rangle_m} \in (0, 1].
  \label{eq:internal_drift}
\end{equation}
Let $d^{(k)}_{i,t}=\log\pi_k(y_{i,t}\mid x_i,y_{i,<t})-\log\pi_{k-1}(y_{i,t}\mid x_i,y_{i,<t})$ denote the single-pass log-probability step on response $i$ (length $L_i$) in prompt group $p\in\mathcal{P}$. Weighting prompts, sibling rollouts $\mathcal{R}(p)$, and valid tokens equally via $w_{i,t}=(|\mathcal{P}|\,|\mathcal{R}(p)|\,L_i)^{-1}$ defines per-prompt and batch marginal realization $M_k$ as well as cumulative advantage-aligned work $W_k$,
\begin{equation}
  M_{k,p} = \frac{1}{|\mathcal{R}(p)|}\sum_{i\in\mathcal{R}(p)} \frac{1}{L_i}\sum_{t=1}^{L_i} A_{0,i,t}\, d^{(k)}_{i,t},
  \qquad
  M_k = \frac{1}{|\mathcal{P}|}\sum_{p\in\mathcal{P}} M_{k,p},
  \qquad
  W_k = \sum_{j=1}^{k} M_j = \left\langle A_0,\, \Delta^{(k)} \right\rangle_w,
  \label{eq:marginal}
\end{equation}
with within-batch 95\% lower confidence bound $\operatorname{LCB}(M_k) = M_k - 1.96\,\hat{\sigma}_k/\sqrt{|\mathcal{P}|}$, where $\hat{\sigma}_k$ is the sample standard deviation of $\{M_{k,p}\}_{p\in\mathcal{P}}$.

Global fixed depth and online per-batch depth test distinct properties. Let $V_n^{(3)}$ be the downstream long-horizon value of running E3 on batch $n$, and let $z_n\in\{0,1\}$ be a selector decision with mean rate $\mathbb{E}[z_n]=b$. Raising fixed depth from $D=2$ to $D=3$ helps whenever $\mathbb{E}[V_n^{(3)}] > 0$. An online selector beats an equal-rate random rule only when $\mathbb{E}[(z_n - b)V_n^{(3)}] = \operatorname{Cov}(z_n, V_n^{(3)}) > 0$. Because $M_2$ measures local movement on batch $n$ rather than future trajectory value $V_n^{(3)}$, secular growth in the scale of $M_2$ across training can make a static token price spend early and starve late steps, driving $\operatorname{Cov}(z_n, V_n^{(3)})$ negative even when $M_2$ strongly ranks local $M_3$.

\section{Experimental setup}
\label{sec:setup}

\subsection{Training protocol and compared methods}
\label{sec:h20_setup}

We distill Qwen3-30B-A3B-Instruct-2507~\citep{qwen2025qwen3a3b2507} into Qwen3-1.7B-Base~\citep{yang2025qwen3} on DAPO-Math-17k~\citep{yu2025dapo} using HybridFlow/veRL~\citep{sheng2024hybridflow} and vLLM~\citep{kwon2023vllm}. Each main training run executes 300 steps with 32 prompts per batch, eight rollouts per prompt, a 4096-token response limit, four PPO mini-batch updates per pass at learning rate $10^{-6}$, and checkpoints every 50 steps. Adaptive arms solve Equation~\ref{eq:alpha} over a 32-point log grid $\mathcal{G}\subset[0.01,1]$ with $\delta=0.02$ and $\tau=-0.5$.

\begin{table}[t]
\caption{Summary of the six 300-step training arms, grouped into one-pass baselines, optional third-pass allocation controls under a matched token cap, and deterministic SAVOR schedules.}
\label{tab:arms}
\begin{center}
\small
\setlength{\tabcolsep}{3.5pt}
\begin{tabular*}{\linewidth}{@{\extracolsep{\fill}}lllll}
\toprule
System & Target weight $\alpha$ & Actor passes & Third-pass rule & Normalization \\
\midrule
OPD & Raw teacher ($\alpha=1$) & E1 ($D=1$) & None & Budget-preserving \\
TOP-D & Fixed $\alpha=0.3$ & E1 ($D=1$) & None & Per-prompt z-score \\
\midrule
TPMR & Adaptive $\alpha^\star$ (Eq.~\ref{eq:alpha}) & E1+E2+(E3) & Token-priced $M_2$ (134/300) & Budget-preserving \\
SAVOR-Random & Adaptive $\alpha^\star$ (Eq.~\ref{eq:alpha}) & E1+E2+(E3) & Random $p=0.5$ (144/300) & Budget-preserving \\
\midrule
SAVOR-2 & Adaptive $\alpha^\star$ (Eq.~\ref{eq:alpha}) & E1+E2 ($D=2$) & None & Budget-preserving \\
SAVOR-3 & Adaptive $\alpha^\star$ (Eq.~\ref{eq:alpha}) & E1+E2+E3 ($D=3$) & Every batch (300/300) & Budget-preserving \\
\bottomrule
\end{tabular*}
\end{center}
\end{table}

Table~\ref{tab:arms} organizes the six main runs into three groups. Raw OPD ($\alpha=1$) and TOP-D ($\alpha=0.3$, chosen from the recommended range of~\citet{xie2026topd} and retaining its native per-prompt z-score normalization, which outperforms budget-preserving normalization at one pass) are one-pass baselines. SAVOR-2 and SAVOR-3 are deterministic two-pass and three-pass schedules that isolate pass depth under identical target calibration and normalization. TPMR and SAVOR-Random diagnose online per-batch depth allocation under a shared third-pass token cap and safety gates, where TPMR triggers E3 when $\operatorname{LCB}(M_2)$ per million tokens clears a 50-step pilot price and SAVOR-Random triggers E3 with probability $0.5$ without reading $M_2$.

Alongside the 300-step main study, we run a controlled 50-step ablation suite under identical models, data, and optimizer hyperparameters with shared budget-preserving normalization across all arms (never pooled with 300-step endpoints). This suite isolates constant $\alpha=0.3$ versus adaptive $\alpha^\star$, frozen $A_0$ versus post-E1 residual or alignment updates, and pass depth $D\in\{1,2,3\}$. Cluster hardware, serving layout, normalization formulas, and online controller gates are detailed in Appendix~\ref{app:normalization}.

\subsection{Benchmarks and evaluation}
\label{sec:eval}

We evaluate checkpoints on five mathematical reasoning benchmarks~\citep{hendrycks2021math,lightman2024verify}, comprising AIME 2024, AIME 2025, and AIME 2026 (30 Olympiad problems each), AMC 2023 (40 problems), and MATH-500 (500 problems). Following standard stochastic reasoning evaluation, we sample 32 responses per problem on AIME and AMC and 4 on MATH-500 with an exact-answer verifier, reporting per-benchmark accuracy, the three-AIME mean, and the unweighted five-benchmark macro average. Main endpoint comparisons always use step 300 without cherry-picking intermediate checkpoints.

To quantify finite-benchmark sampling uncertainty, we compute 95\% percentile intervals from 10,000 paired problem-bootstrap resamples within each dataset. Cost accounting records cumulative teacher-scored tokens, cumulative actor-processed tokens across executed passes, and wall-clock training time on identical hardware nodes. Dataset identifiers, decoding hyperparameters, and multi-seed decoding stability checks are reported in Appendix~\ref{app:trajectories}.

\section{Main empirical results}
\label{sec:results}

\subsection{Accuracy at matched teacher budgets}
\label{sec:main}

\begin{figure}[t]
\begin{center}
\includegraphics[width=\linewidth]{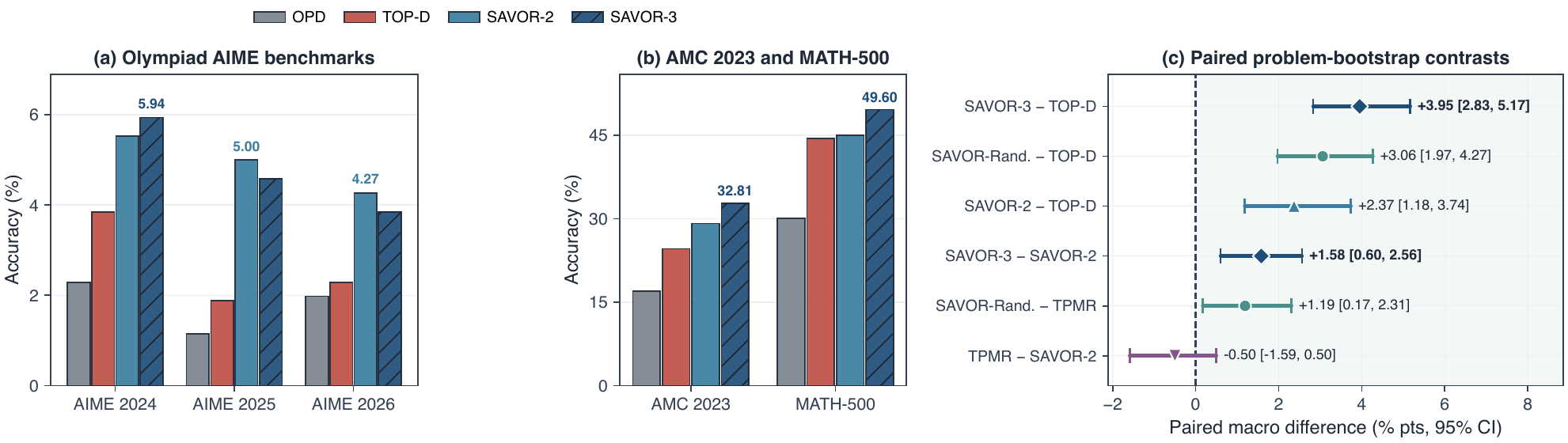}
\end{center}
\caption{Per-benchmark step-300 accuracy and paired uncertainty. (a) Exact-match accuracy on Olympiad benchmarks AIME 2024, AIME 2025, and AIME 2026. (b) Accuracy on AMC 2023 and MATH-500. (c) Forest plot of 95\% paired problem-bootstrap intervals for five-benchmark macro accuracy differences across key policy contrasts.}
\label{fig:forest}
\end{figure}

\paragraph{Deeper realization drives large gains without extra teacher queries.}
One-pass distillation leaves open whether student accuracy is limited by how many prefixes the teacher scores or by how thoroughly the student optimizes prefixes already scored. Table~\ref{tab:main} and Figure~\ref{fig:forest} isolate realization depth by holding teacher-scored volume matched within 1\% across TOP-D and all four adaptive arms, ensuring that endpoint differences come from how paid batches are calibrated and reused rather than from broader teacher coverage. Bounding negative log-ratio spikes in one-pass TOP-D lifts raw OPD by $+4.91$ macro points (10.50\% to 15.41\%). Optimizing each acquired batch deeper inside a batchwise KL envelope adds a further $+2.37$ points in SAVOR-2 (17.78\%) and $+3.95$ points in SAVOR-3 (19.36\%), recovering four fifths of the entire jump from raw OPD to TOP-D without a single extra teacher forward pass. Every adaptive multi-pass schedule outperforms TOP-D on all five benchmarks, and the paired bootstrap intervals in Figure~\ref{fig:forest}c separate both the gains over TOP-D and the $+1.58$-point gain from SAVOR-2 to SAVOR-3 from benchmark sampling noise.

\begin{table}[t]
\caption{Main step-300 results across five mathematical reasoning benchmarks. Accuracy ($\uparrow$) is generation-level exact-match percentage. E3 counts batches executing a third pass, and Tch.\ M and Act.\ M report cumulative teacher-scored and actor-processed tokens in millions. Bold marks the column best.}
\label{tab:main}
\begin{center}
\small
\setlength{\tabcolsep}{3.4pt}
\begin{tabular*}{\linewidth}{@{\extracolsep{\fill}}lrrrrrrrrr}
\toprule
System & E3 & Tch.\ M & Act.\ M & AIME24 $\uparrow$ & AIME25 $\uparrow$ & AIME26 $\uparrow$ & AMC23 $\uparrow$ & MATH $\uparrow$ & Macro $\uparrow$ \\
\midrule
OPD & 0 & 306.3 & 306.3 & 2.29 & 1.15 & 1.98 & 17.03 & 30.05 & 10.50 \\
TOP-D & 0 & 292.5 & 292.5 & 3.85 & 1.88 & 2.29 & 24.61 & 44.40 & 15.41 \\
\midrule
TPMR & 134 & 292.6 & 713.7 & 4.06 & 3.44 & 3.85 & 29.84 & 45.20 & 17.28 \\
SAVOR-Random & 144 & 292.0 & 723.5 & \textbf{6.35} & 3.85 & \textbf{4.69} & 30.00 & 47.45 & 18.47 \\
\midrule
SAVOR-2 & 0 & 293.0 & 586.0 & 5.52 & \textbf{5.00} & 4.27 & 29.14 & 44.95 & 17.78 \\
SAVOR-3 & 300 & 291.4 & 874.1 & 5.94 & 4.58 & 3.85 & \textbf{32.81} & \textbf{49.60} & \textbf{19.36} \\
\bottomrule
\end{tabular*}
\end{center}
\end{table}

\paragraph{Where a third pass helps depends on student task support.}
Splitting results by benchmark difficulty exposes a clear regime boundary between Olympiad competition sets and mid-difficulty reasoning tasks. On the three AIME benchmarks (Figure~\ref{fig:forest}a), where the 1.7B student solves under 6\% of rollouts, all adaptive multi-pass arms roughly double TOP-D (4.79\% to 4.96\% three-AIME mean versus 2.67\%) yet overlap one another within problem-sampling uncertainty. When few student rollouts reach valid Olympiad derivations, wide teacher-student divergence forces conservative $\alpha^\star$ weights and two passes already capture the coarse structural corrections available on each batch. On AMC 2023 and MATH-500 (Figure~\ref{fig:forest}b), where the student already places substantial probability mass on complete solution paths, narrower divergence admits stronger $\alpha^\star$ targets and sharper within-prompt advantage contrast, allowing a third pass to refine multi-step algebraic execution and lift SAVOR-3 over SAVOR-2 by $+3.67$ and $+4.65$ points. Meanwhile, spending 134 selective third passes under pilot-calibrated token pricing (TPMR, 17.28\%) trails both cap-matched SAVOR-Random (18.47\%) and two-pass SAVOR-2, showing that when extra passes occur over training matters as much as their total count. Full pairwise bootstrap intervals are detailed in Appendix~\ref{app:bootstrap}.

\subsection{Training dynamics across checkpoints}
\label{sec:trajectories}

\begin{figure}[t]
\begin{center}
\includegraphics[width=0.96\linewidth]{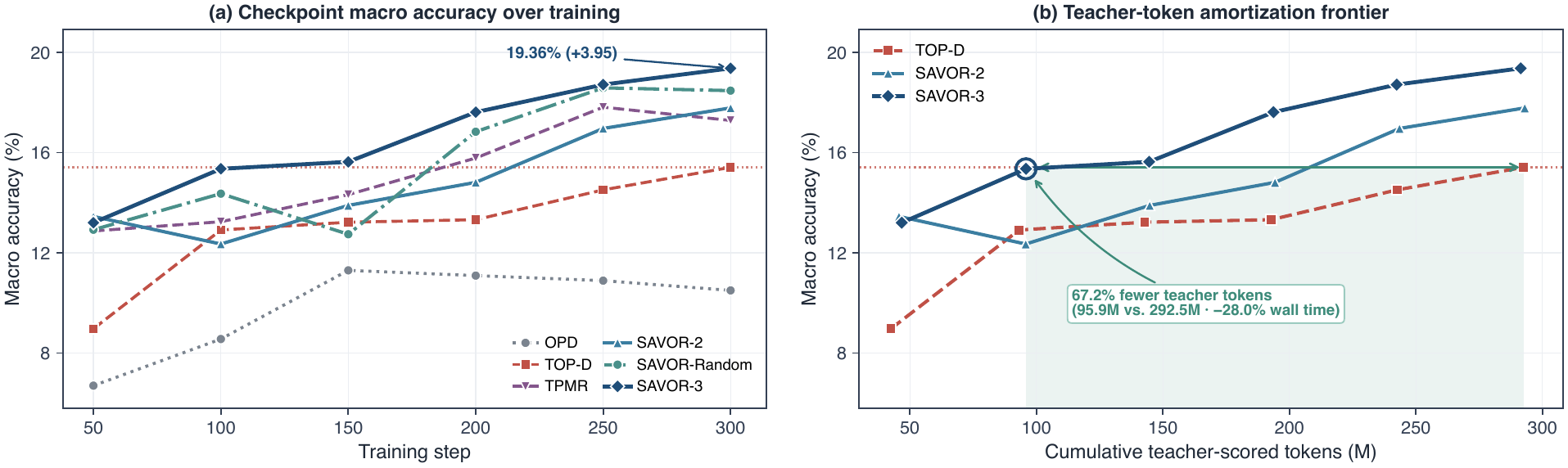}
\end{center}
\caption{Training dynamics and teacher-token efficiency across 300 steps. (a) Five-benchmark macro accuracy across all six 50-step checkpoints. (b) Macro accuracy against cumulative teacher-scored tokens, highlighting step-100 SAVOR-3 (circled) matching 300-step TOP-D with two thirds fewer teacher tokens.}
\label{fig:training_efficiency}
\end{figure}

Figure~\ref{fig:training_efficiency}a contrasts three optimization regimes across the six 50-step checkpoints. Raw OPD climbs through step 150 (11.30\%) and then regresses steadily, as unclipped negative log-ratio penalties on off-teacher prefixes destabilize updates once student entropy shifts. Constant proximal weighting in TOP-D restores monotonic improvement from step 50 onward, but progresses slowly through step 200 because a single clipped pass extracts limited movement per batch. By throttling $\alpha^\star$ on wide-gap early batches and relaxing it as the student aligns, SAVOR-3 at step 50 (13.20\%) already matches step-150 TOP-D (13.22\%) and leads every arm from step 100 through the end of training. TPMR tracks ahead of SAVOR-2 through step 250 before slipping over the final 50 steps once early spending exhausts most of its third-pass allowance. Complete checkpoint trajectories are reported in Appendix~\ref{app:trajectories}.

\subsection{Substituting student depth for costly teacher queries}
\label{sec:amortization}

In synchronous reasoning distillation, step wall time is dominated by multi-thousand-token rollout generation and 30B teacher forward passes rather than 1.7B actor backpropagation. Under a fixed 300-batch horizon, processing $2\times$ and $3\times$ as many actor tokens raises total training time in SAVOR-2 and SAVOR-3 to $1.65\times$ and $2.19\times$ TOP-D. Comparing runs at matched actor optimization instead (287.8M versus 292.5M actor tokens in Figure~\ref{fig:training_efficiency}b and Table~\ref{tab:milestone}) tests whether deeper student reuse can substitute directly for fresh teacher queries. At step 100, SAVOR-3 performs the same total actor gradient work as 300-step TOP-D while skipping two thirds of rollout generation and teacher scoring, matching TOP-D's final accuracy (15.35\% versus 15.41\%) with 67.2\% fewer teacher-scored tokens and 28.0\% less training time ($0.72\times$). By step 150, SAVOR-3 surpasses TOP-D's endpoint (15.63\%) at half the teacher budget. Teacher breadth and student pass depth therefore trade off smoothly along a shared cost-accuracy frontier, letting practitioners either stop early to save teacher calls and wall time or invest extra passes on a fixed prompt budget for higher peak accuracy.

\begin{table}[t]
\caption{Teacher-signal and wall-time amortization against 300-step TOP-D. Tch.\ M and Act.\ M report cumulative teacher-scored and actor-processed tokens in millions, and Rel.\ time reports wall-clock training time relative to 300-step TOP-D on identical hardware.}
\label{tab:milestone}
\begin{center}
\small
\begin{tabular}{lrrrrr}
\toprule
System & Step & Tch.\ M $\downarrow$ & Act.\ M & Rel.\ time $\downarrow$ & Macro (\%) $\uparrow$ \\
\midrule
SAVOR-3 & 50  & \textbf{46.8}  & 140.4 & \textbf{0.35$\times$} & 13.20 \\
SAVOR-3 (matched actor) & 100 & 95.9 & 287.8 & 0.72$\times$ & 15.35 \\
SAVOR-3 & 150 & 144.6 & 433.7 & 1.09$\times$ & \textbf{15.63} \\
\midrule
TOP-D (endpoint) & 300 & 292.5 & 292.5 & 1.00$\times$ & 15.41 \\
\bottomrule
\end{tabular}
\end{center}
\end{table}

\section{Mechanism analysis and diagnostics}
\label{sec:analysis}

\subsection{Why single-pass tuning misleads multi-pass reuse}
\label{sec:feasible}

\begin{figure}[t]
\begin{center}
\includegraphics[width=\linewidth]{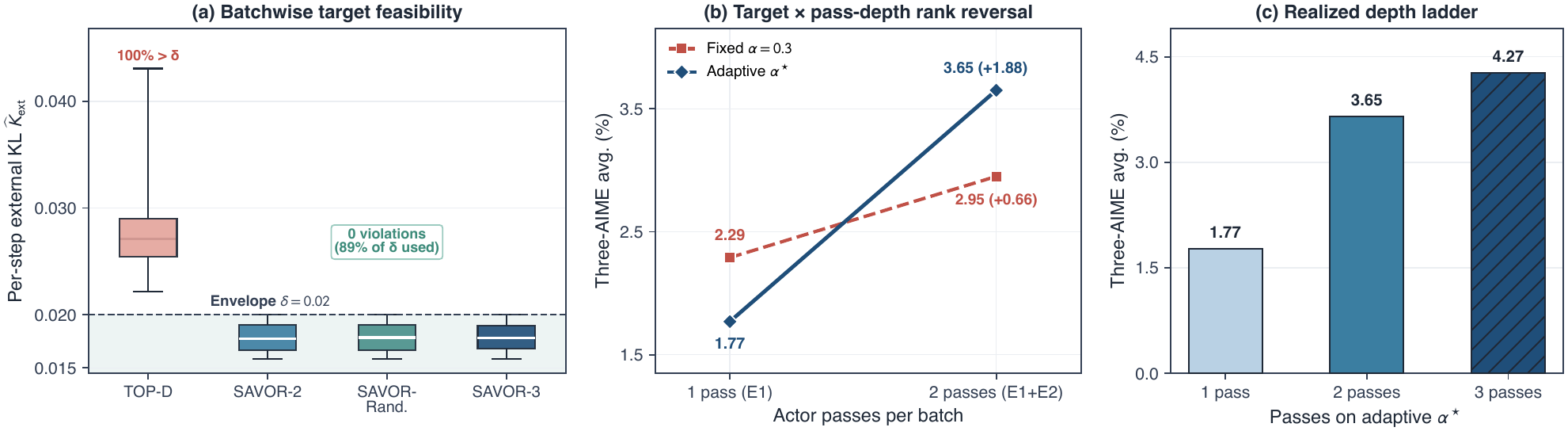}
\end{center}
\caption{Mechanism analysis of target feasibility and pass depth. (a) Per-step external KL divergence $\widehat{K}_{\mathrm{ext}}$ across 300 training steps (interquartile boxes, medians, and $[\text{min}, \text{max}]$ whiskers). (b) Controlled $2\times 2$ ablation at step 50 under shared normalization, showing the rank reversal from $D=1$ to $D=2$. (c) Controlled step-50 pass-depth ladder under adaptive $\alpha^\star$.}
\label{fig:mechanism}
\end{figure}

\paragraph{Constant weights break batchwise trust regions while adaptive targets track the envelope.}
Figure~\ref{fig:mechanism}a explains why no fixed mixture weight stays well-calibrated across training. Under constant $\alpha=0.3$ in TOP-D, external reverse KL $\widehat{K}_{\mathrm{ext}}$ exceeds $\delta=0.02$ on every step of the 300-step run, averaging 138\% of the envelope and reaching 0.0431 on wide-gap batches, whereas lowering constant $\alpha$ to $0.2$ in calibration traces avoids violations at the cost of leaving 28\% of $\delta$ idle. Across all adaptive runs, Equation~\ref{eq:alpha} varies $\alpha^\star$ nearly threefold across batches (0.0928 to 0.2626), recording zero violations while using 89\% of $\delta$ on average.

\paragraph{Out-of-envelope targets win at one pass but stall on repetition, and mid-step edits add noise.}
Our controlled 50-step $2\times 2$ ablation under identical normalization reveals why standard one-pass sweeps miss multi-pass gains (Figure~\ref{fig:mechanism}b). Evaluated after a single pass (E1), out-of-envelope $\alpha=0.3$ beats feasible $\alpha^\star$ on three-AIME accuracy (2.29\% versus 1.77\%) because an aggressive step pushes probability mass rapidly on easy tokens while PPO clipping limits single-epoch damage on outlier prefixes. Repeating a second pass (E1+E2) against $\alpha=0.3$ drives token ratios into the clipping ceiling and adds only $+0.66$ points (2.95\%), whereas feasible $\alpha^\star$ keeps updates inside an active trust region and compounds by $+1.88$ points to 3.65\%, reversing the one-pass ranking. Raising depth to three passes lifts three-AIME accuracy further to 4.27\% (Figure~\ref{fig:mechanism}c). Holding $\alpha^\star$ and $D=2$ fixed in the same suite also shows why $q_{\alpha^\star}$ must act as a stationary batchwise anchor. Recomputing remaining gaps ($g_{\alpha^\star,t} - \Delta^{(1)}_t$) or update-alignment weights after E1 lowers accuracy from 3.65\% to between 2.88\% and 3.37\%, because mini-batch stochasticity within E1 causes single-pass log-probability changes to overshoot or flip sign on roughly 57\% of valid token positions. All nine controlled ablation arms are summarized in Appendix~\ref{app:a100}.

\subsection{Fixed global depth versus online token-priced allocation}
\label{sec:allocation}

Under shared target calibration and normalization in the 300-step study, increasing the share of batches executing E3 from 0\% (SAVOR-2) to 48\% (SAVOR-Random) and 100\% (SAVOR-3) produces a monotonic accuracy ladder (Table~\ref{tab:main} and Figure~\ref{fig:forest}c), confirming that the mean long-horizon value $\mathbb{E}[V_n^{(3)}]$ of a third pass is positive at zero extra teacher cost. Within SAVOR-3, second-pass realization $\operatorname{LCB}(M_2)$ also ranks within-batch third-pass realization $\operatorname{LCB}(M_3)$ with Spearman $\rho=0.996$ ($\rho\in[0.913,0.964]$ per 50-step window). Yet pricing E3 against a 50-step pilot threshold in TPMR loses by 1.19 macro points to uninformative coin flips in SAVOR-Random (Figure~\ref{fig:forest}c) and fails to beat SAVOR-2.

\begin{figure}[t]
\begin{center}
\includegraphics[width=0.96\linewidth]{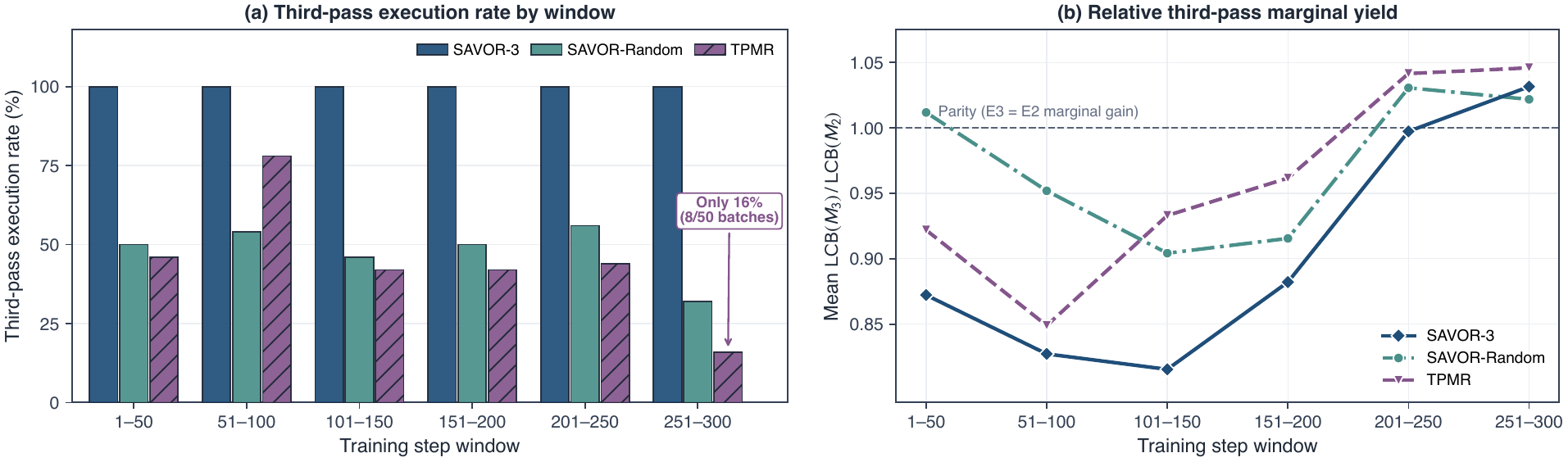}
\end{center}
\caption{Windowed analysis of third-pass allocation across 50-step training intervals. (a) Share of batches executing E3 per window, showing early over-spending and late starvation under TPMR. (b) Mean ratio $\operatorname{LCB}(M_3)/\operatorname{LCB}(M_2)$ over batches executing E3 in each window, rising above parity over the final 100 steps.}
\label{fig:windows}
\end{figure}

Figure~\ref{fig:windows} traces this reversal to cross-stage scale non-stationarity. As the student aligns with the teacher and Equation~\ref{eq:alpha} admits larger $\alpha^\star$ weights, the magnitude of $M_2$ expands roughly tenfold over training (Figure~\ref{fig:diagnostics}a). TPMR rejects 27 early batches during steps 1 to 50, then treats mid-run batches as apparent bargains once $M_2$ crosses its static pilot price, executing E3 on 78\% of steps 51 to 100 (Figure~\ref{fig:windows}a). That front-loaded spend briefly lifts TPMR past SAVOR-2 at step 150, but triggers pace and budget caps that leave only 8 third passes (16\%) for steps 251 to 300, right when narrowing teacher-student gaps push relative third-pass yield $\operatorname{LCB}(M_3)/\operatorname{LCB}(M_2)$ from 0.87 above parity to 1.03 (Figure~\ref{fig:windows}b). Static pilot pricing thus spends early and starves late, driving $\operatorname{Cov}(z_n, V_n^{(3)})$ negative in Section~\ref{sec:depth_theory} whereas fixed global depth $D\in\{2,3\}$ captures late-stage gains without pilot calibration or extra forward passes.

\subsection{Testing objective exhaustion and off-policy drift across passes}
\label{sec:diagnostics}

\begin{figure}[t]
\begin{center}
\includegraphics[width=\linewidth]{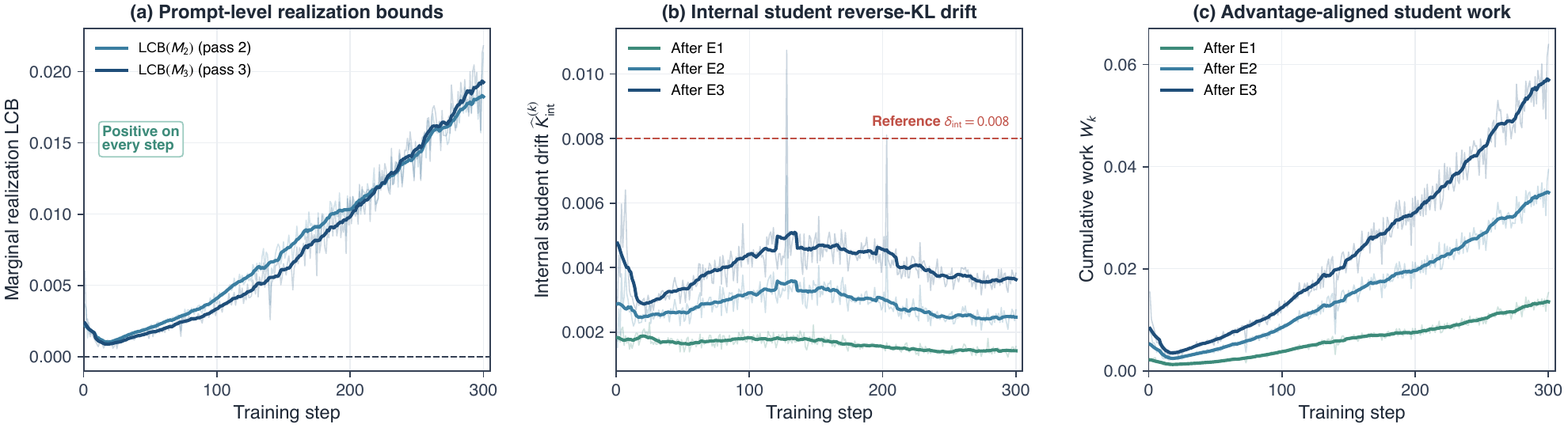}
\end{center}
\caption{Pass-level diagnostics across all 300 steps of SAVOR-3, plotting raw step values (translucent) and 15-step rolling means (solid). (a) Prompt-level bounds $\operatorname{LCB}(M_2)$ and $\operatorname{LCB}(M_3)$ stay positive at every step. (b) Internal reverse-KL drift $\widehat{K}_{\mathrm{int}}^{(k)}$ stays well below the external envelope. (c) Cumulative advantage-aligned work $W_k$ increases cleanly from E1 to E3.}
\label{fig:diagnostics}
\end{figure}

Running three passes per batch raises two concrete optimization questions, namely whether E3 exhausts the frozen surrogate objective and whether cumulative student updates drift too far off-policy from $\pi_0$. Figure~\ref{fig:diagnostics} rules out both failure modes across all 300 steps of SAVOR-3. First, $\operatorname{LCB}(M_2)$ and $\operatorname{LCB}(M_3)$ remain strictly positive on every step (Figure~\ref{fig:diagnostics}a), and cumulative advantage-aligned work $W_k$ separates cleanly across passes throughout training (Figure~\ref{fig:diagnostics}c). Second, internal reverse-KL drift $\widehat{K}_{\mathrm{int}}^{(k)}$ grows sublinearly to means of 0.0016, 0.0029, and 0.0040 after E1, E2, and E3 (Figure~\ref{fig:diagnostics}b), remaining roughly one fifth of the external envelope $\delta=0.02$ and crossing the 0.008 reference line on only 2 of 300 steps. Unclipped importance-sampling effective sample size stays at 0.981 after E2 and 0.971 after E3, confirming that importance ratios remain well-conditioned without replay buffers or teacher refreshes.

\subsection{Token-level entropy, conflict reduction, and problem-level gains}
\label{sec:token_cases}

\begin{figure}[t]
\begin{center}
\includegraphics[width=\linewidth]{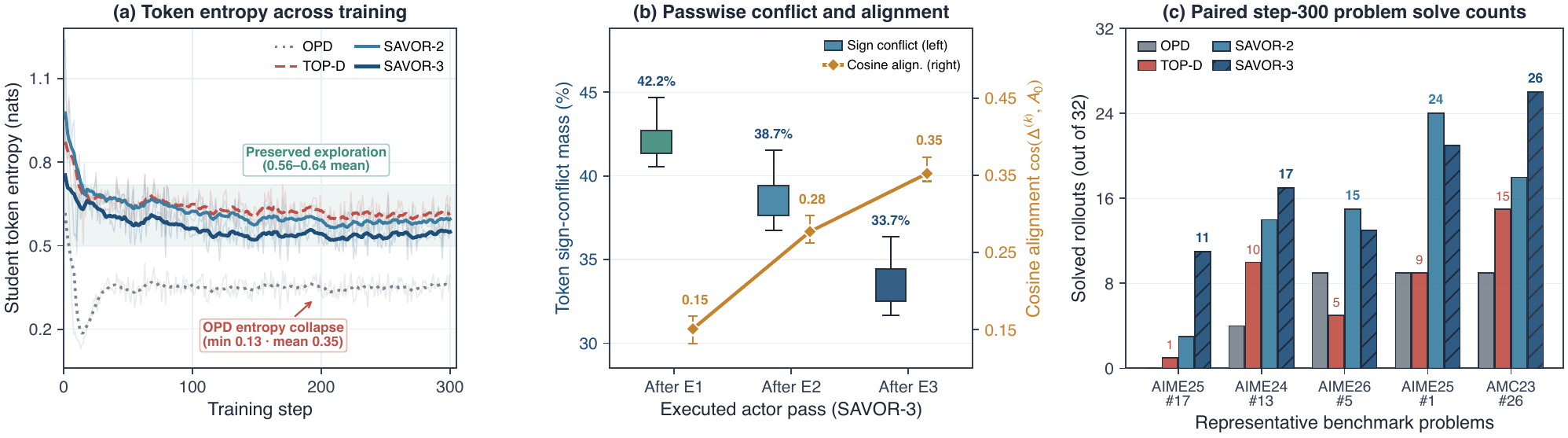}
\end{center}
\caption{Token-level dynamics and concrete problem-level gains across 300 steps. (a) Student token entropy over training (raw translucent and 15-step rolling means), contrasting entropy collapse under raw OPD against preserved exploration under bounded proximal targets. (b) Passwise $|A_{0,t}|$-weighted token sign-conflict rate (5th to 95th percentile whiskers and interquartile boxes, left axis) and token-weighted cosine alignment $\cos(\Delta^{(k)}, A_0)$ (mean and interquartile bars, right axis) across SAVOR-3 passes. (c) Paired step-300 solve counts out of 32 generations on representative multi-step benchmark problems.}
\label{fig:token_dynamics}
\end{figure}

Figure~\ref{fig:token_dynamics} connects macro benchmark gains to token-level behavior during training and concrete problem outcomes at evaluation. Under raw OPD, unbounded negative log-ratio spikes rapidly collapse student token entropy from 1.24 to a minimum of 0.13 nats (0.35 mean in Figure~\ref{fig:token_dynamics}a), explaining why its accuracy peaks mid-run and then degrades once generation diversity vanishes. Bounding the target inside $\delta=0.02$ preserves stable exploration throughout all 300 steps, maintaining mean token entropies of 0.64 in TOP-D, 0.62 in SAVOR-2, and 0.56 in SAVOR-3 even under three gradient passes per batch. Within each SAVOR-3 step, mini-batch gradient interference during E1 leaves 42.2\% of $|A_{0,t}|$-weighted token mass moving opposite in sign to $A_{0,t}$ (Figure~\ref{fig:token_dynamics}b). Sweeping the frozen batch in E2 and E3 resolves that interference monotonically, cutting sign-conflict mass to 38.7\% and 33.7\% while lifting token-weighted cosine alignment $\cos(\Delta^{(k)}, A_0)$ from 0.15 to 0.28 and 0.35. Figure~\ref{fig:token_dynamics}c illustrates how cleaner token updates surface on multi-step competition problems under 32 paired rollouts. On AIME 2025 base divisibility ($17_b \mid 97_b$, \#1) and AIME 2026 acute-angle planar rotation (\#5), calibrated reuse lifts solved rollouts from 9 and 5 out of 32 in TOP-D to 24 and 15 in SAVOR-2, while on harder algebraic chains such as AIME 2025 polynomial divisibility ($n+2 \mid 3(n+3)(n^2+9)$, \#17) and AMC 2023 (\#26), a third pass pushes solved counts from 1 and 15 in TOP-D (3 and 18 in SAVOR-2) up to 11 and 26 in SAVOR-3.

\section{Related work}
\label{sec:related}

\paragraph{On-policy language-model distillation.}
Knowledge distillation compresses a teacher into a student via output logits~\citep{hinton2015distilling}, teacher-sampled sequences~\citep{kim2016sequence}, or hidden representations~\citep{sanh2019distilbert,jiao2020tinybert}. Generalized Knowledge Distillation and OPD move supervision onto student-sampled prefixes to remove exposure mismatch~\citep{agarwal2024opd}. MiniLLM optimizes reverse KL via policy gradients~\citep{gu2024minillm}, DistiLLM mixes skew-KL with off-policy replay~\citep{ko2024distillm}, and OPD+ addresses stop-gradient advantage bias~\citep{zhao2026opdplus}. Recent reasoning methods reweight tokens by entropy or divergence~\citep{jin2026entropyopd,ko2026reopold,xu2026tip}, apply sign-asymmetric objectives~\citep{jia2026aopd}, align hidden states~\citep{yang2026oprd}, or guide rollout tokens~\citep{liu2026tgpo}. While those methods change which signal is collected, SAVOR studies how deeply the student should optimize each batch once scored.

\paragraph{Trust regions and proximal teacher targets.}
KL-constrained policy updates underpin TRPO~\citep{schulman2015trpo}, PPO~\citep{schulman2017ppo}, and constrained policy search~\citep{achiam2017cpo}. In OPD, TOP-D introduces a constant-weight mixture teacher $q_\alpha$~\citep{xie2026topd}, TrOPD restricts updates to reliable regions~\citep{xing2026tropd}, Trust-Region Behavior Blending modifies rollout policies near the student~\citep{plyusov2026trb}, and self-distillation studies examine teacher refresh cadence~\citep{guo2026teachermove}. SAVOR retains the proximal mixture family within a batch, replacing constant $\alpha$ with per-batch constrained maximization inside a sampled K3 and quantile envelope so multi-pass reuse stays feasible across varying teacher-student gaps.

\paragraph{Sample efficiency, staleness, and multi-pass optimization.}
\citet{fu2026oneshot} find strong OPD gains from even a single repeated query, consistent with our observation that one-pass schedules under-use scored batches. Asynchronous and buffered systems such as AsyncOPD~\citep{kang2026asyncopd} and $f$-OPD~\citep{chen2026fopd} control staleness across delayed rollouts, whereas ERPD trains an extreme intermediate policy before a second trust-region distillation stage~\citep{chen2026erpd}. SAVOR is synchronous, needs no cross-step replay buffer or secondary distillation stage, and reuses only the current frozen batch. PPO~\citep{schulman2017ppo}, GRPO~\citep{shao2024deepseekmath}, and TOP-D~\citep{xie2026topd} allow multiple epochs per batch, and empirical RL studies show how strongly such optimizer choices shape on-policy learning~\citep{engstrom2020implementation,andrychowicz2020what}. SAVOR shows that multi-pass OPD requires batchwise target feasibility and uses dual-ledger accounting and allocation controls to compare fixed depth against pilot-calibrated token pricing.

\section{Limitations}
\label{sec:limitations}

Our study has three main scope boundaries. First, our experiments evaluate one student-teacher pair in the Qwen3 family on mathematical reasoning with programmatic verifiers under a 4096-token generation cap, where long chains of thought frequently reach the context limit, and each 300-step trajectory is a single training run assessed via paired problem-bootstrap and repeated-decoding checks rather than multi-seed training runs. Extending SAVOR to cross-family model pairs, open-ended code or dialogue generation, longer context windows, and multi-seed long-horizon runs is an important direction for future work. Second, SAVOR saves teacher queries by running extra student passes. That trade reduces overall training time when rollout generation and large-teacher forward passes dominate step latency, but increases student update cost when the number of rollout batches is held fixed or when actor backpropagation is the primary bottleneck. Third, our online allocation negative result applies specifically to static pilot-calibrated token pricing over $D\in\{2,3\}$. Scale-normalized or rolling-quantile depth controllers, deeper schedules ($D\ge 4$), and adaptive $\delta$ schedules remain open to study.

\section{Conclusion}
\label{sec:conclusion}

Teacher-scored rollouts in on-policy distillation are expensive to collect yet routinely discarded after a single gradient pass. We showed that constant proximal weights cannot track a teacher-student divergence that varies across batches and contracts over training, causing single-pass tuning to favor aggressive targets whose gains saturate under reuse. SAVOR resolves this mismatch by selecting the strongest envelope-feasible mixture weight for each batch, freezing the scored trajectories and token advantages, and optimizing the student across multiple PPO passes with zero extra rollouts or teacher calls. On mathematical reasoning benchmarks, SAVOR substantially improves accuracy at matched teacher budgets and reaches baseline final accuracy using a fraction of the teacher queries. Together with our analysis showing how non-stationary update scales undermine static token pricing, these findings point to batchwise target calibration paired with fixed global pass depth as a simple and effective recipe for synchronous on-policy distillation.

\subsubsection*{Broader Impact Statement}
This work reduces the teacher-scoring and hardware cost of distilling mathematical reasoning from large teacher models into compact students, lowering training energy use and enabling capable local models. At the same time, extracting multiple student passes per teacher-scored batch without an external ground-truth verifier can more tightly transfer teacher biases, flawed reasoning habits, or dual-use capabilities into unmonitored open weights. Practitioners applying SAVOR in general domains should curate training prompts, audit teacher generations before multi-pass distillation, and run safety and bias evaluations prior to deployment. All experiments use public open-weight models and symbolic mathematics datasets containing no personally identifiable information or human-subject data.

\bibliography{savor}
\bibliographystyle{tmlr}

\appendix

\section{Checkpoint trajectories, evaluation details, and decoding repeats}
\label{app:trajectories}

We evaluate checkpoints on AIME 2024 (\texttt{Maxwell-Jia/AIME\_2024}, 30 problems), AIME 2025 (\texttt{evalscope/aime25}, 30 problems), AIME 2026 (\texttt{evalscope/aime26}, 30 problems), the 2023 AMC subset (\texttt{AI-MO/aimo-validation-amc}, 40 problems), and MATH-500 (\texttt{HuggingFaceH4/MATH-500}, 500 problems), decoding at temperature 1.0, top-$p$ 0.7, maximum length 4096, and primary seed 20260727. Problem-paired bootstrap resampling is supported by strong cross-arm problem-level accuracy correlations ($r\in[0.73,1.00]$ versus $0.01$ after random permutation). Table~\ref{tab:trajectory} lists five-benchmark macro accuracy at every predeclared 50-step checkpoint of the 300-step study.

\begin{table}[!ht]
\caption{Five-benchmark macro accuracy (\%) across all saved 50-step checkpoints of the 300-step study.}
\label{tab:trajectory}
\begin{center}
\small
\begin{tabular}{lrrrrrr}
\toprule
System & Step 50 & Step 100 & Step 150 & Step 200 & Step 250 & Step 300 \\
\midrule
OPD & 6.70 & 8.56 & 11.30 & 11.09 & 10.89 & 10.50 \\
TOP-D & 8.97 & 12.91 & 13.22 & 13.32 & 14.51 & 15.41 \\
\midrule
TPMR & 12.87 & 13.24 & 14.32 & 15.78 & 17.81 & 17.28 \\
SAVOR-Random & 12.92 & 14.36 & 12.74 & 16.83 & 18.58 & 18.47 \\
\midrule
SAVOR-2 & \textbf{13.45} & 12.35 & 13.89 & 14.81 & 16.96 & 17.78 \\
SAVOR-3 & 13.20 & \textbf{15.35} & \textbf{15.63} & \textbf{17.61} & \textbf{18.71} & \textbf{19.36} \\
\bottomrule
\end{tabular}
\end{center}
\end{table}

Table~\ref{tab:repeats} evaluates conditional decoding variability at steps 50, 100, and 150 using auxiliary decoding seeds distinct from the primary seed (three independent repeats per cell, five for TPMR). At step 150, within-checkpoint range is at most 0.51 macro points across all six arms and at most 0.27 points across the five non-TPMR arms.

\begin{table}[!ht]
\caption{Repeated-decoding five-benchmark macro mean and $[\text{min}, \text{max}]$ range (\%) across auxiliary decoding seeds at intermediate checkpoints.}
\label{tab:repeats}
\begin{center}
\small
\begin{tabular}{lccc}
\toprule
System & Step 50 & Step 100 & Step 150 \\
\midrule
OPD & 6.81 $[6.73, 6.94]$ & 8.41 $[8.27, 8.65]$ & 11.17 $[11.08, 11.24]$ \\
TOP-D & 9.30 $[9.19, 9.45]$ & 12.88 $[12.76, 12.93]$ & 13.89 $[13.81, 13.97]$ \\
\midrule
TPMR & 12.67 $[12.50, 12.82]$ & 13.36 $[13.09, 13.60]$ & 14.39 $[14.20, 14.71]$ \\
SAVOR-Random & 12.71 $[12.65, 12.77]$ & 14.39 $[14.26, 14.50]$ & 12.66 $[12.62, 12.72]$ \\
\midrule
SAVOR-2 & 12.99 $[12.73, 13.13]$ & 12.51 $[12.35, 12.69]$ & 13.81 $[13.68, 13.91]$ \\
SAVOR-3 & 12.99 $[12.92, 13.07]$ & 15.47 $[15.30, 15.63]$ & 16.07 $[15.94, 16.21]$ \\
\bottomrule
\end{tabular}
\end{center}
\end{table}

\section{Paired problem bootstrap intervals}
\label{app:bootstrap}

Table~\ref{tab:bootstrap} lists 95\% percentile intervals from 10,000 paired problem resamples at step 300 of the main study. Every pairwise difference resamples the identical problem indices for both arms within each benchmark before computing macro or three-AIME means.

\begin{table}[!ht]
\caption{Standalone and pairwise 95\% problem-bootstrap intervals (\%) at step 300.}
\label{tab:bootstrap}
\begin{center}
\small
\begin{tabular}{lr}
\toprule
Quantity & 95\% bootstrap interval \\
\midrule
Macro accuracy, TOP-D & $[13.04, 17.87]$ \\
Macro accuracy, TPMR & $[14.53, 20.16]$ \\
Macro accuracy, SAVOR-Random & $[15.57, 21.60]$ \\
Macro accuracy, SAVOR-2 & $[14.74, 20.99]$ \\
Macro accuracy, SAVOR-3 & $[16.39, 22.45]$ \\
\midrule
Macro gain, SAVOR-2 minus TOP-D & $[1.18, 3.74]$ \\
Macro gain, SAVOR-Random minus TOP-D & $[1.97, 4.27]$ \\
Macro gain, SAVOR-3 minus TOP-D & $[2.83, 5.17]$ \\
Macro gain, SAVOR-3 minus SAVOR-2 & $[0.60, 2.56]$ \\
Macro gain, SAVOR-Random minus SAVOR-2 & $[-0.19, 1.59]$ \\
Macro gain, SAVOR-Random minus TPMR & $[0.17, 2.31]$ \\
Macro gain, TPMR minus SAVOR-2 & $[-1.59, 0.50]$ \\
\midrule
AIME avg.\ gain, SAVOR-2 minus TOP-D & $[0.69, 4.17]$ \\
AIME avg.\ gain, SAVOR-3 minus TOP-D & $[0.94, 3.58]$ \\
AIME avg.\ gain, SAVOR-3 minus SAVOR-2 & $[-1.35, 1.08]$ \\
\bottomrule
\end{tabular}
\end{center}
\end{table}

\section{Infrastructure, advantage normalization, and online controller details}
\label{app:normalization}

Each 300-step main trajectory ran on one dedicated node of eight NVIDIA H20 GPUs using seed 42, prompt limit 1024, rollout temperature 1.0, top-$p$ 1.0, and PPO mini-batch size 8 prompts. Four GPUs executed student rollouts and actor updates (per-GPU token cap 8192) and four served the teacher with tensor parallelism four. Multiplying elapsed step wall time by the eight occupied GPUs yields step-300 totals of 136.8 GPU-hours for OPD, 131.5 for TOP-D, 271.0 for TPMR, 250.5 for SAVOR-Random, 216.7 for SAVOR-2, and 288.1 for SAVOR-3, and intermediate SAVOR-3 totals of 46.2, 94.7, and 142.8 GPU-hours at steps 50, 100, and 150. Across the four adaptive 300-step runs, 1,199 of 1,200 steps produced complete KL logs (excluding one split stdout line at SAVOR-2 step 275), with $\widehat{K}_{\mathrm{ext}}\in[0.0159,0.0200]$ and zero violations. Roughly 89.1\% of audited SAVOR training rollouts and over 92\% in TOP-D and OPD reached the 4096-token cap.

Let $\mathcal{T}(p)$ denote the set of valid token positions $(i,t)$ across all sibling rollouts $i\in\mathcal{R}(p)$ for prompt $p$. The per-prompt z-score normalization of TOP-D~\citep{xie2026topd} computes
\begin{equation}
  A^{\mathrm{PZ}}_{i,t} = \frac{R_{i,t} - \operatorname{mean}_{(u,v)\in\mathcal{T}(p)}(R_{u,v})}{\operatorname{std}_{(u,v)\in\mathcal{T}(p)}(R_{u,v}) + \epsilon}.
\end{equation}
SAVOR retains prompt-group centering while dividing by a slow cross-batch scale that preserves relative return spread across prompts. Letting $d_b$ be the mean across prompt groups in batch $b$ of $\operatorname{median}_{(i,t)\in\mathcal{T}(p)}|R_{i,t} - \operatorname{mean}_{(u,v)\in\mathcal{T}(p)}(R_{u,v})|$ with $s_0=1$, budget-preserving normalization updates
\begin{equation}
  s_b = 0.99\,s_{b-1} + 0.01\,d_b,
  \qquad
  A^{\mathrm{BP}}_{i,t} = \frac{R_{i,t} - \operatorname{mean}_{(u,v)\in\mathcal{T}(p)}(R_{u,v})}{\max(s_b, 10^{-6})}.
\end{equation}
OPD and all SAVOR and TPMR arms use $A^{\mathrm{BP}}_{i,t}$, while TOP-D uses $A^{\mathrm{PZ}}_{i,t}$.

In the online allocation diagnostic, TPMR runs E1 and E2 on every batch $n\in\{1,\ldots,300\}$, re-forwards $\pi_2$ to compute $g_n = \max(\operatorname{LCB}(M_{2,n}), 0)$, and evaluates utility $u_n = 10^6 g_n / \max(c_n, 1)$ for single-pass valid token count $c_n$. TPMR executes E3 when $u_n \ge \lambda = 0.00134707$ (calibrated solely on pre-E3 statistics from a 50-step SAVOR-3 pilot) and safety, budget, and pace gates hold. Safety requires at least eight prompts, finite statistics, $\widehat{K}_{\mathrm{int}}^{(2)} < 0.012$, and $\mathrm{ESS}^{(2)} > 0.50$ computed on importance ratios clipped at 100, plus a watchdog entering a 10-step cooldown if two consecutive $M_3$ LCBs are nonpositive or the trailing median of up to eight $\operatorname{LCB}(M_3)/\operatorname{LCB}(M_2)$ values falls below 0.50. The hard third-pass cap is 140,361,966 tokens (90\% price-selected and 10\% reserved for 0.1-probability probes, yielding 131 price-selected and 3 probe executions), with cumulative spend at step $n$ paced below $\min(1, n/300 + 0.10)$ of each account cap. SAVOR-Random shares the 140.4M cap, pace ceiling, and finite/$\widehat{K}_{\mathrm{int}}^{(2)}$/100-clipped $\mathrm{ESS}^{(2)}$ safety gates, triggering E3 with probability 0.5 without reading $M_2$ or running the $M_3$ watchdog.

\section{Controlled 50-step mechanism and signal-reuse ablations}
\label{app:a100}

Table~\ref{tab:a100} compiles all controlled 50-step ablation arms discussed in Section~\ref{sec:setup} and Section~\ref{sec:analysis}. Every arm ran on one node of eight NVIDIA A100-SXM4-80GB GPUs with Qwen3-1.7B-Base, Qwen3-30B-A3B-Instruct-2507, DAPO-Math-17k, seed 42, batch size 32, eight rollouts per prompt, learning rate $10^{-6}$, and 32-generation exact-verifier evaluation on AIME 2024, 2025, and 2026. Additional 12-step KL stress probes across seeds 42, 1, and 7 confirmed that fixed $\alpha=0.3$ violated $\delta=0.02$ on all 12 steps per seed (averaging 154\% of $\delta$) while adaptive $\alpha^\star$ recorded zero violations. Longer calibration traces on the same cluster recorded zero violations at constant $\alpha=0.2$ (averaging 72\% of $\delta$ over 150 steps) and at adaptive $\alpha^\star$ (averaging 89\% of $\delta$ over 200 steps).

\begin{table}[H]
\caption{Controlled step-50 mechanism and signal-reuse ablations. AIME avg.\ (\%) averages exact-match generation accuracy across AIME 2024, 2025, and 2026.}
\label{tab:a100}
\begin{center}
\small
\setlength{\tabcolsep}{3.5pt}
\begin{tabular*}{\linewidth}{@{\extracolsep{\fill}}llllr}
\toprule
Target $\alpha$ & Normalization & Signal reuse rule & Passes & AIME avg.\ (\%) $\uparrow$ \\
\midrule
Fixed $\alpha=0.3$ & TOP-D z-score & Single pass $A_0$ & E1 ($D=1$) & 3.02 \\
Fixed $\alpha=0.3$ & Budget-preserving & Single pass $A_0$ & E1 ($D=1$) & 2.29 \\
Fixed $\alpha=0.3$ & Budget-preserving & Frozen $A_0$ & E1+E2 ($D=2$) & 2.95 \\
\midrule
Adaptive $\alpha^\star$ & Budget-preserving & Single pass $A_0$ & E1 ($D=1$) & 1.77 \\
Adaptive $\alpha^\star$ & Budget-preserving & Remaining-gap correction & E1+E2 ($D=2$) & 2.88 to 3.09 \\
Adaptive $\alpha^\star$ & Budget-preserving & Update-alignment weighting & E1+E2 ($D=2$) & 3.37 \\
Adaptive $\alpha^\star$ & Budget-preserving & Frozen $A_0$ (SAVOR-2) & E1+E2 ($D=2$) & 3.65 \\
Adaptive $\alpha^\star$ & Budget-preserving & Token-priced optional E3 & E1+E2+(E3) & 3.51 \\
Adaptive $\alpha^\star$ & Budget-preserving & Frozen $A_0$ (SAVOR-3) & E1+E2+E3 ($D=3$) & \textbf{4.27} \\
\bottomrule
\end{tabular*}
\end{center}
\end{table}

\end{document}